\pdfoutput=1
\documentclass[11pt]{article}

\usepackage[]{acl}
\usepackage{placeins}
\usepackage{times}
\usepackage{enumitem}
\usepackage{latexsym}
\usepackage{multirow}
\usepackage{algorithm}
\usepackage{algpseudocode}
\usepackage{xcolor}
\usepackage{graphicx}
\usepackage{tabularx}
\usepackage{graphicx}  % Required for \resizebox
\usepackage{tabularx}  % Allows flexible table column widths
\usepackage{booktabs}  % Improves table formatting
\usepackage{multirow}  % Enables multirow cells
\newcommand{\email}[1]{\ttfamily\begin{tabular}{@{} c @{}}#1\end{tabular}}
\usepackage[T1]{fontenc}
\usepackage[utf8]{inputenc}

\usepackage{microtype}

\usepackage{inconsolata}

\usepackage{todonotes}

\title{DETQUS: Decomposition-Enhanced Transformers\\ for QUery-focused Summarization}

\author{
Yasir Khan, 
Xinlei Wu, 
Sangpil Youm,
Justin Ho, 
Aryaan Shaikh, \\ 
\textbf{Jairo Garciga}, 
\textbf{Rohan Sharma},
\textbf{Bonnie J. Dorr} \\
        University of Florida, Gainesville, Florida\\
        \email{\{y.khan, x.wu, youms, justinho, am.shaikh, \\ jgarciga, rohansharma1,  bonniejdorr\}@ufl.edu}
}
\begin{document}
\maketitle
\begin{abstract}

% Query-focused tabular summarization is a new task in the area of table-to-text that maps tabular data along with user queries to a synthesized summary response. We introduce DETQUS (Decomposition-Enhanced Transformers for QUery-focused Summarization), a system designed to address the token limit constraints of transformer models while enhancing summarization accuracy. DETQUS leverages tabular decomposition combined with a fine-tuned encoder-decoder model to improve task performance. Tabular decomposition employs a large language model to decompose and shrink tables, ensuring that only relevant columns remain based on a user query. This method reduces input size while preserving essential information, enabling DETQUS to process larger tables more effectively. Our approach yields an improved ROUGE-L score of 0.4437, outperforming the previous state-of-the-art REFACTOR approach (ROUGE-L: 0.422) on the Omnitab model. Furthermore, we demonstrate that DETQUS provides a promising and efficient alternative to more complex architectures, offering a structured method to improve query-focused tabular summarization.

Query-focused tabular summarization is an emerging task in table-to-text generation that synthesizes a summary response from tabular data based on user queries. Traditional transformer-based approaches face challenges due to token limitations and  the complexity of reasoning over large tables. To address these challenges, we introduce \textbf{DETQUS} (\textbf{D}ecomposition-\textbf{E}nhanced \textbf{T}ransformers for \textbf{QU}ery-focused \textbf{S}ummarization), a system designed to improve summarization accuracy by leveraging tabular decomposition alongside a fine-tuned encoder-decoder model. DETQUS employs a large language model to selectively reduce table size, retaining only query-relevant columns while preserving essential information. This strategy enables more efficient processing of large tables and enhances summary quality. Our approach, equipped with table-based QA model Omnitab,  achieves a ROUGE-L score of 0.4437, outperforming the previous state-of-the-art REFACTOR model (ROUGE-L: 0.422). These results highlight DETQUS as a scalable and effective solution for query-focused tabular summarization, offering a structured alternative to more complex architectures.

\end{abstract}
%This task is challenging due to the inherent token limit in traditional state-of-the-art transformer techniques and %the complex reasoning required to handle user queries related to a table. To address these challenges, we explore the %use of tabular decomposition combined with a fine-tuned encoder-decoder model. Tabular decomposition is a technique %that utilizes a large language model to decompose and shrink a table so that, based on a user query, only the relevant %columns are left. This technique effectively reduces the input size while preserving essential information, enabling %our method to handle larger tables while also improving the summarization accuracy. Our approach yields an improved %ROUGE-L score of 0.4437, surpassing the previous state-of-the-art REFACTOR approach, which scores 0.422 on the Omnitab %model. Furthermore, we illustrate that tabular decomposition offers a promising and efficient alternative to more %complex architectures previously explored for this task. 

% \bonnieshort{DETQUS is not properly introduced anywhere on page 1, not in the abstract, not in the intro?  IS THIS INTENTIONAL  I noticed in your response, you said you'd introduce DETQUS in the abstract (and even had quoted text).  Also, Please don't say "We propose DETQUS" since it's an implemented system, say "We introduce DETQUS...." -- review your responses, not all are implemented; some are less important actually, but this one is REALLY important.}

\section{Introduction}
\label{sec:intro}

Tabular data has become increasingly prevalent in our society, with nearly all businesses employing it to store crucial information. As the amount of collected data increases over time, this has created the need for techniques and systems that allow individuals to analyze and create insights about their data. Automatic summarization is an area of research that investigates methods to %create 
glean insights from natural language data using various techniques \cite{SurveySummary}. 

The importance of analyzing tabular data combined with the existing summarization research has given rise to a new subset of summarization:
query-focused tabular summarization. Query-focused tabular summarization refers to the task of extracting key points and context from large tables based on a user-provided query \citep{zhao2023qtsumm}. 

\begin{figure}
    % \begin{center}
    % \text{Table 1: Communication with Extraterrestrial Intelligence}
    \includegraphics[width=1\linewidth]{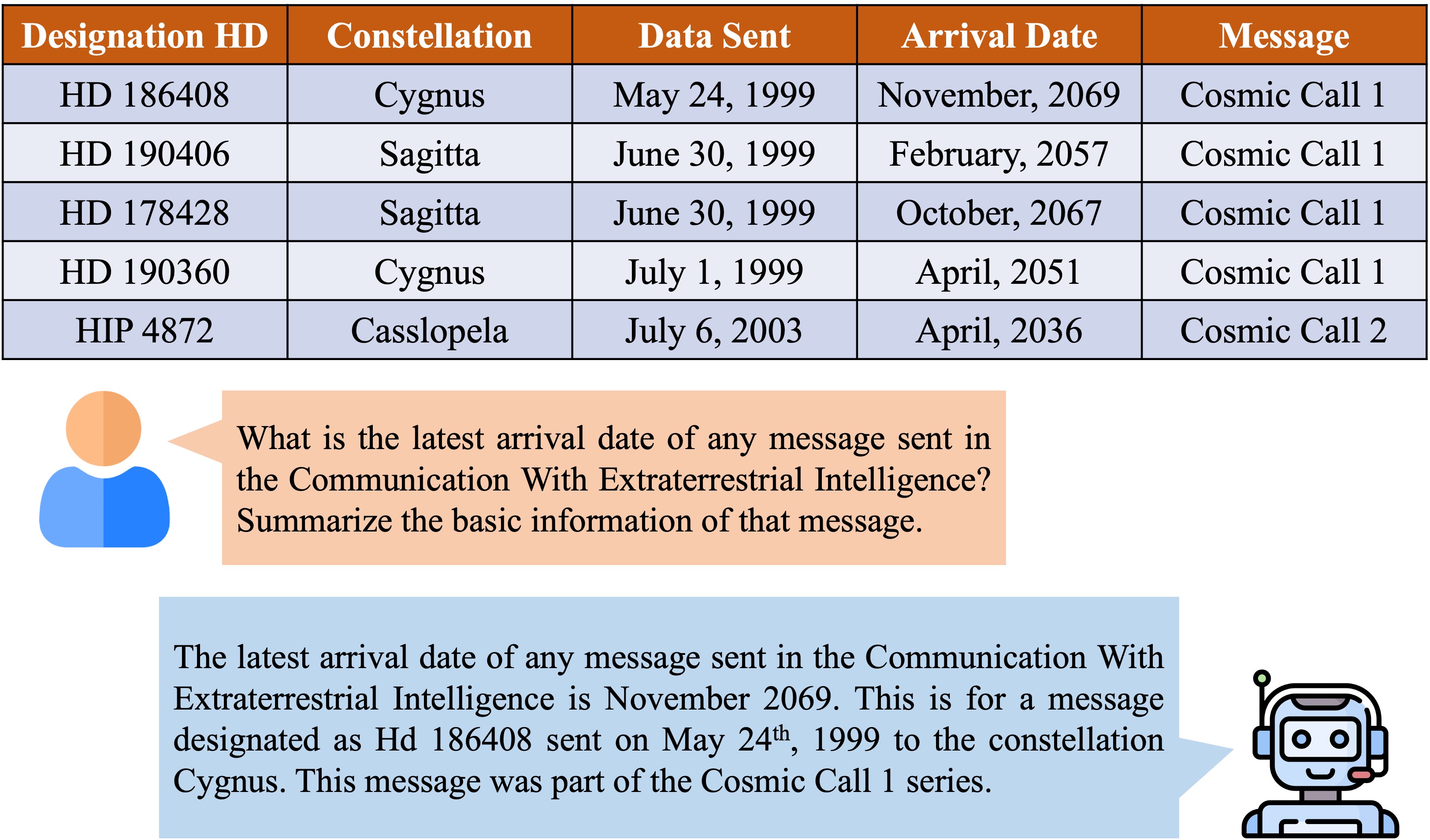}
    % \includegraphics[width=1\linewidth]{latex//Images/example1.png}
    % \end{center}
    % \includegraphics[width=0.9\linewidth]{latex//Images/egx1.png}
    % \begin{flushright}
    % \includegraphics[width=0.9\linewidth]{latex//Images/egx2.png}
    \vspace*{-.1in}
    \caption{Query-focused table summarization with QTSUMM, generating a summary from the query: 
    %Query-focused table summarization using QTSUMM table. %\cite{zhao2023qtsumm}. 
    %A summary is generated based on the decomposition of the query 
    \textit{``What is the latest ... intelligence?''}} 
    \label{fig:example}
    % \end{flushright}
    \vspace*{-.1in}
\end{figure}

% To address this challenge, we introduce DETQUS (Decomposition-Enhanced Transformers for QUery-focused Summarization), which improves query-based table summarization by leveraging tabular decomposition. DETQUS efficiently restructures table to mitigate token length constraints in transformer-based models while preserving relevant content. Our system processes complex queries by dynamically decomposing tables into more manageable structures, allowing neural models to generate accurate and concise summaries.

Considering the query-focused table summarization system depicted in Figure~\ref{fig:example}, when a user inquires about ``arrival date'' and ``basic information,'' our model evaluates the user's requirements in conjunction with a table lookup and then generates a summary based on these constraints.

% \begin{figure}
%     % \begin{center}
%     % \text{Table 1: Communication with Extraterrestrial Intelligence}
%     \includegraphics[width=1\linewidth]{latex/Images/Modified_Figure1.jpeg}
%     % \includegraphics[width=1\linewidth]{latex//Images/example1.png}
%     % \end{center}
%     % \includegraphics[width=0.9\linewidth]{latex//Images/egx1.png}
%     % \begin{flushright}
%     % \includegraphics[width=0.9\linewidth]{latex//Images/egx2.png}
%     \vspace*{-.1in}
%     \caption{Query-focused table summarization with QTSUMM, generating a summary from the query: 
%     %Query-focused table summarization using QTSUMM table. %\cite{zhao2023qtsumm}. 
%     %A summary is generated based on the decomposition of the query 
%     \textit{``What is the latest ... intelligence?''}} 
%     \label{fig:example}
%     % \end{flushright}
%     \vspace*{-.3in}
% \end{figure}
% \bonnieshort{A rei}

We consider using neural techniques, particularly transformer-based models, and identify several associated challenges. The first challenge is the %innate 
inherent token capacity limitation of transformer models. When tables are converted to text and tokenized, they can exceed the input length of the transformer model, causing performance degradation due to truncation from the overly long sequence length. Another challenge is processing larger tables and complex queries, which require models to reason across numerous columns, establish meaningful connections, and synthesize coherent summaries from the extracted information. Although large language models (LLMs) %sufficiently large LLMs 
exhibit emergent reasoning abilities and greater token capacity, their reasoning abilities still falls short
%have been found to still 
% to be worse than that of people, limiting their performance in this area 
of human performance~\cite{wei2022emergent, davis2023reasoning}.
% \sangpilshort[]{We need to introduce DETQUS here, please add paragraph or few sentences that introduces DETQUS. such as ``We introduce DETQUS (Decomposition...)}

To address these limitations, we introduce \textbf{DETQUS} (\textbf{D}ecomposition-\textbf{E}nhanced \textbf{T}ransformers for \textbf{QU}ery-focused \textbf{S}ummarization), a novel system designed to enhance query-based table summarization through tabular decomposition. DETQUS dynamically restructures tables into smaller, more relevant forms based on the provided query, effectively mitigating token length constraints while preserving critical content. The principle behind this technique is that the model only accesses data relevant to improving performance. This helps mitigate the token limitation of many attention-based methods and the complexity of reasoning over large tables and complex queries.
% \justin[]{So as far as I am aware, ye2023 is the first to use LLMs as decomposers to improve tabular qa models. We should elaborate here a little bit more on how we build upon their work. I.e we apply it to this task that it hasn't been used in before, we expand the algorithm or process by doing this, etc. Pretty much I want to make sure the reviewer can distinguish our work from this prior work.  I made some modifications but I don't think that's enough - Xinlei }

Our work builds on the foundational research of~\citet{ye2023large}, introducing tabular decomposition based on the given query. We extend this approach to query-focused table summarization, making several key contributions. Our main contribution is developing DETQUS, a system that leverages tabular decomposition to enhance query-based table summarization. DETQUS outperforms prior baselines while offering a more structured and interpretable approach. We evaluate our method across multiple transformer-based models, demonstrating improvements in summarization quality through a combination of fine-tuned neural architectures and optimized decomposition techniques.

% \sangpilshort[]{I added this section to replace bullet point, feel free to edit this if needed}  
% \xinleishort[]{I modified the contribution paragraph to emphasize DETQUS}

% \begin{itemize}[noitemsep, topsep=0pt, leftmargin=*]
%   \item A tabular decomposition method for query-based table summarization that provides performance improvements beyond a prior baseline using a far simpler and more explainable architecture.
%   \item An analysis of the impact of tabular decomposition on a wide range of transformers-based models.
%   \item An error analysis of query-based table summarization using the tabular decomposition method and current weaknesses of the system.
%    \item A human evaluation study rating the output of the tabular decomposition when combined with certain models.
% \end{itemize}

%We introduce related work in 
Section~\ref{sec:relatedWork} presents related work, followed by a description of
%. We then present 
our system architecture for tabular summarization in Section~\ref{sec:architect}. We present
%, followed by its 
data and experiments in Section~\ref{sec:DataExperiment}. %, with
%. Results and analyses are shown in 
% results and analyses in Section~\ref{sec:ResultsAnalyases}. Our model yields a ROUGE-L score of 0.4385, surpassing 0.410 
% achieved by the REFACTOR approach on text generation models. 
% \cite{zhao2023qtsumm}. \bonnieshort{So now this is REFACTOR, before it was QTSUMM. I tried to fix the wording in the Intro and Fig 1 caption; someone please check what I did and cross-check it with the edits here as well.}
Finally, we
%then discuss the insight 
discuss insights gleaned from our study and provide conclusions and future directions.

\section{Related Work}
\label{sec:relatedWork}

This section outlines the task of query-focused tabular summarization, highlighting how it differs from other table-to-text tasks. We review prior approaches to these tasks and their limitations, comparing them to our method. We also describe the QTSUMM dataset which we use for our evaluation. 
% \bonnieshort{I added missing punctuation here. Also, are you sure you implemented your promised changes in this section?  I see something about Llama3 context window--is that supposed to be in here?  Maybe it is dropped appropriately, but I'm pointing it out because I'm wondering if some IMPORTANT comments have not been integrated...?  And also, this section is still super long (you promised to shorten it), so you should have room for integrating those comments.}

\subsection{Explanation of the Task}

Query-focused tabular summarization is a specific type of table-to-text generation that combines elements of table question answering (QA) and generic table summarization \cite{zhao2023qtsumm}. Unlike tabular QA, which focuses on extracting specific facts from a table based on a query, query-focused tabular summarization aims to generate a coherent summary that addresses the user’s query by integrating relevant information from the table. This task is distinct from generic table summarization, which generates summaries based solely on the tabular input without regard to a specific query.

\subsection{Prior Approaches to Tabular QA and Summarization}

Previous research in tabular question answering primarily focuses on models that can extract relevant facts from tables in response to specific queries. These models typically rely on sophisticated parsing techniques and neural networks to understand and retrieve the 
correct information. For example, the method developed by
\citet{chen2021open} extracts facts from tables given a particular query. While this approach provides a high degree of user control, its focus on fact extraction does not allow for generating insightful summaries or interpretations beyond the presented data.

Recent advancements in table-based question answering
(QA) highlight the effectiveness of pretraining models with both natural and synthetic data to improve few-shot learning scenarios. \citet{jiang2022omnitab} introduce OmniTab, a model pre-trained using natural sentences paired with tables and synthetic questions derived from SQL queries. This dual approach enhances the model's ability to align natural language with tabular data and perform complex reasoning tasks. 
% \bonnieshort{I fixed a major orphan/widow issue here, three of them, by using the word "show" instead of "suggest" and "balances" instead of "can balance".}
% The OmniTab model demonstrates significant improvements in few-shot and full settings on the WikiTableQuestions benchmark, establishing new state-of-the-art performance. Their findings show that integrating both natural and synthetic data effectively balances alignment and reasoning capabilities, providing a robust foundation for table-based QA systems.
OmniTab achieves state-of-the-art performance on the WikiTableQuestions benchmark, demonstrating that integrating both data types balances alignment and reasoning.

Generic table summarization involves generating summaries based solely on the tabular input without specific queries, as seen in the work of \citet{lebret-etal-2016-neural}. This method, while straightforward, lacks user control over the summary content. To address this limitation, single-sentence table-to-text tasks are introduced by \citet{chen-etal-2020-logical}, which provide models with tabular data and specific sequences describing rows and columns. Although this offers some level of control, it still requires manual input for regions of interest, limiting its flexibility as a system.

\subsection{Advances in Query-focused Text Summarization}
Recent advancements in query-focused text summarization have improved the 
efficiency and relevance of summaries for specific user queries. \citet{rahman2015survey} demonstrate extractive techniques 
%where the goal is to distill 
aimed at distilling information for a specific query.
%from one or more documents. 
\citet{xu2022document} introduce latent queries to bridge the gap between explicit user queries and implicit information, leveraging latent semantic analysis for deeper query understanding.
Although these approaches
%, while primarily focused on 
primarily target textual data, 
they provide a foundation 
for developing techniques applicable to tabular data.

\subsection{Advances in Table Summarization Techniques}

Previous developments in table summarization techniques focus on improving the ability of models to generate accurate and relevant summaries from structured data. Techniques such as TAPAS \cite{herzig2020tapas} and TAPEX \cite{liu2021tapex} demonstrate significant progress in enhancing reasoning capabilities over tabular data. These models employ pre-trained transformers designed specifically to handle structured information, achieving strong performance on table-based reasoning and question-answering tasks.

Despite these advancements, there are notable limitations, particularly in handling larger or more complex tables. Models like TAPAS and TAPEX often struggle with scalability and suffer from token limitations, as they may not efficiently process tables with a large number of columns or rows. To address these issues, REASTAP \cite{zhao2022reastap} introduces table reasoning skills during pre-training, improving performance on specific tasks like table-based question answering. However, even REASTAP encounters difficulties related to token limitations and integrating unstructured data with structured inputs.

Our approach introduces a novel tabular decomposition technique using a large language model (LLM) to decompose tables based on the user query. This method addresses token capacity constraints by reducing the table to its most relevant columns and rows, allowing the model to handle larger datasets more effectively while retaining essential information for generating accurate summaries.

Although few-shot learning is widely used to improve model performance across various NLP tasks, recent research in query-focused summarization, such as \citet{zhao2023qtsumm}, reveals that it does not always yield significant improvements for table-based tasks. Informed by these results, we decide not to implement few-shot learning in our study and instead focus on alternative methods, such as tabular decomposition, to enhance performance.
% \bonnieshort{Orphan/widow alert: Please make sure all your lines are at least 50\% full; if not, you'll need to ask a LLM to help you rephrase and tighten the last 1-2 sentences of the paragraph, without losing content, to get rid of orphan/widows}

%However, despite these advancements, there are inherent limitations within these models that our architecture aims to address. For instance, TAPAS and TAPEX focus heavily on reasoning over structured data but may struggle with scalability when dealing with exceptionally large or complex tables. REASTAP improves on this by incorporating reasoning capabilities tailored for specific tasks, but it still encounters challenges related to token limitations and the integration of unstructured data with structured tabular inputs.

%Our approach differs in that it introduces a novel table decomposition technique that utilizes an LLM and a user's query to compress a table to only the relevant rows and columns. This decomposition process reduces the input size while retaining essential information, enabling our model to handle larger tables more effectively. Furthermore, by fine-tuning transformer-based models, we enhance the model's capability to generate coherent summaries that are both contextually relevant and accurate.

\subsection{QTSUMM Dataset}

The QTSUMM dataset,\footnote{QTSUMM dataset is publicly available, see here, \url{https://huggingface.co/datasets/yale-nlp/QTSumm}} introduced by \cite{zhao2023qtsumm}, is a comprehensive resource for training and evaluating models on query-focused tabular summarization. 
% This dataset comprises 7,111 query-summary pairs over 2,934 tables, providing a diverse range of topics and scenarios. The tables are curated from LOGICNLG and ToTTo datasets, ensuring a wide variety of data points and user-customized summaries based on specific queries. This diversity makes QTSUMM an ideal dataset for developing and testing models designed to handle query-specific summarization tasks.
This dataset includes tables and query-summary pairs, with a single table potentially having more than one query-summary pair. The tables are scraped from Wikipedia and contain diverse topics. In total, the QTSUMM dataset consists of 7,111 query-summary pairs over 2,934 tables \cite{zhao2023qtsumm}. 
% \sangpilshort[]{we need to provide license and intended Use for used data using footnote}

This dataset is curated from the LOGICNLG \citep{chen-etal-2020-logical} and ToTTo \citep{parikh-etal-2020-totto} tables derived from Wikipedia. Next, tables that are excessively large or small, possess only string-type columns or have hierarchical structures are filtered out from this pool. It is worth noting that topically, the tables used in this dataset are diverse, ranging from sports to scientific literature. This provides a wide domain of tables, queries, and summaries to evaluate models on.

\subsection{Our Methodological Approaches Derived from Previous Works}

Our research methodology addresses the limitations of traditional query-focused table summarization by focusing on a few key innovations. Unlike traditional methods that use either the entire table or manually selected regions, our approach utilizes LLMs to decompose tables based on the query. This method accommodates token-limit constraints while retaining critical information necessary for generating accurate summaries. 

For tabular decomposition, we implement a strategy that uses LLMs to perform table decomposition. For queries that the LLM understands and can confidently identify relevant columns, a precise decomposition is performed in which a table is compressed to the necessary rows and columns. However, for more complex queries where the LLM is less certain, the decomposition is conservative, retaining all potentially relevant columns. This adaptive approach strikes a balance between providing focused information for straightforward queries and maintaining comprehensive context for more complex ones, thus mitigating the risk of information loss and potential model hallucinations.

\begin{figure}
    % \begin{center}
    % \text{Table 1: Communication with Extraterrestrial Intelligence}
    \includegraphics[width=0.95\linewidth]{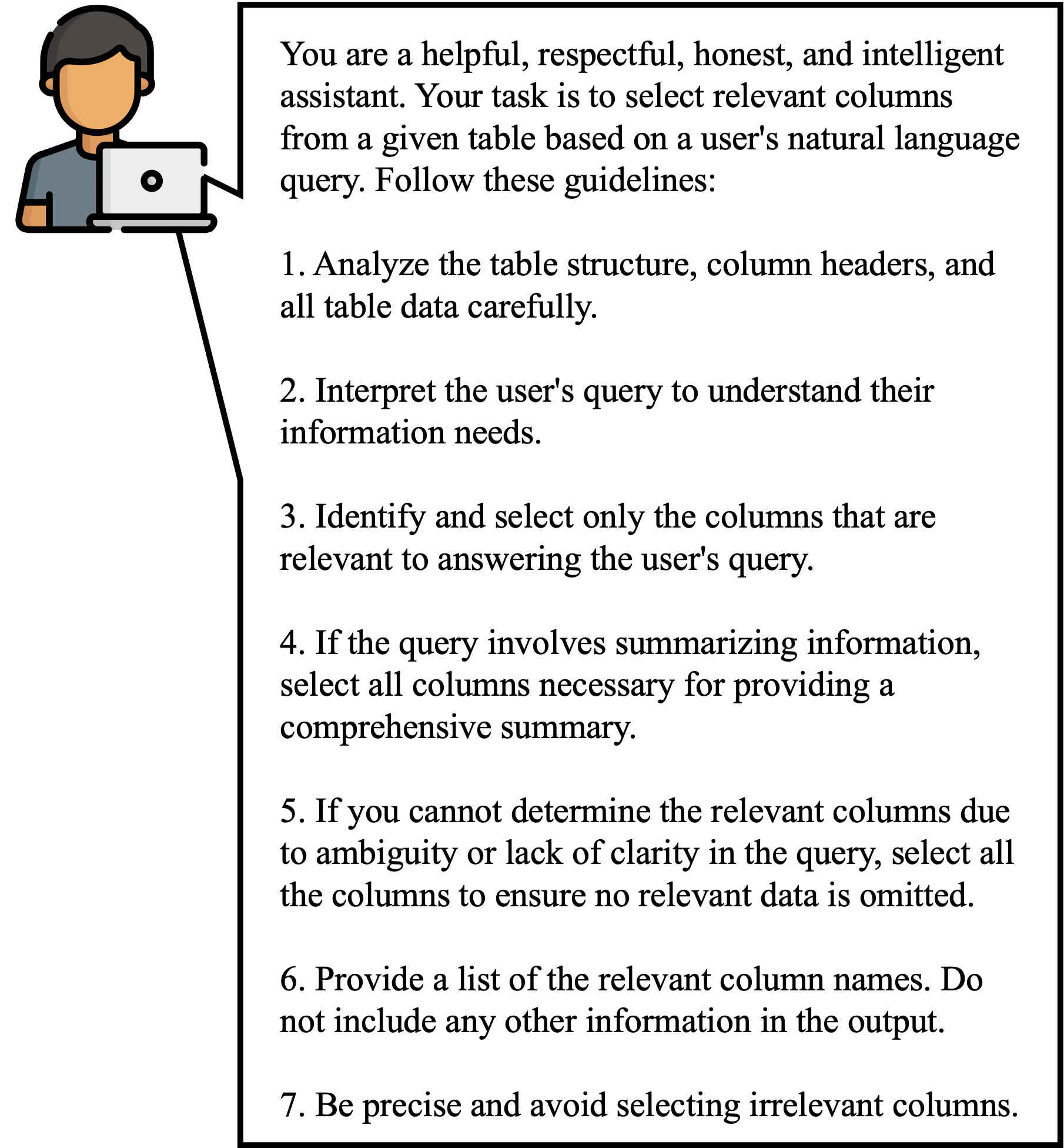}
    % \includegraphics[width=1\linewidth]{latex//Images/example1.png}
    % \end{center}
    % \includegraphics[width=0.9\linewidth]{latex//Images/egx1.png}
    % \begin{flushright}
    % \includegraphics[width=0.9\linewidth]{latex//Images/egx2.png}
    % \vspace*{-.1in}
    \caption{Prompt for converting the table to markdown format for LLM (Llama3-70b)} 
    \label{fig:prompt}
    % \end{flushright}
    \vspace*{-.1in}
\end{figure}

\section{System Architecture for Tabular Summarization}
\label{sec:architect}
This section outlines the architecture of our approach to the tabular decomposition. We describe the table decomposition process and present the algorithmic details of our implementation.

\subsection{Table Decomposition}
Building on \citet{ye2023large}'s work, our approach uses a large language model (LLM) to guide the tabular decomposition, operating in two stages: compression and table-building. In the compression stage, we first convert the table to markdown format for LLM processing. Using Llama3-70b with a tailored prompt (see Figure~\ref{fig:prompt}), we identify columns most relevant to the user's query. These columns are then used in the table-building stage to construct the final decomposed table. 
\subsection{Algorithms for Table Decomposition}

Algorithm~\ref{alg:Table_decomposition} outlines a process for decomposing a table into smaller relevant %smaller 
tables based on a user query, using LLMs to guide the decomposition. The algorithm runs through procedure, MAIN PROCESS (lines 28-33), utilizing % and it utilizes 
two important functions: table decomposition and creating a decomposed table. 

The TABLE\_DECOMPOSITION function first converts the input table to markdown format and constructs a prompt combining the user's question, table content, and specific instructions for column selection. This prompt is then sent to the LLM API for processing. The CREATE\_DECOMPOSED\_TABLE function takes the LLM's response and uses it to build a new, focused table. It searches for column names mentioned in the LLM's output and creates a subset table containing only those columns. Importantly, if no relevant columns are identified in the LLM's response, the function defaults to using all columns from the original table, ensuring robustness against ambiguous queries or unclear LLM responses. 

\begin{algorithm}
\scriptsize
\caption{Table Decomposition and Creation}
\label{alg:Table_decomposition}
    \begin{algorithmic}[1]
    
\Function{table\_decomposition}{\textit{table}, \textit{question}, \textit{title}}
    \State Markdown format $\gets$ \textit{table}
    \State Construct prompt:
    \State \hspace{\algorithmicindent} Add instructions for column relevance
    \State \hspace{\algorithmicindent} Add the given \textit{question}
    \State \hspace{\algorithmicindent} Add the table in markdown format
    \State \hspace{\algorithmicindent} Add the table \textit{title}
    \State Send prompt to LLM API
    \State Retrieve and return the API response
\EndFunction

\Function{create\_decomposed\_table}{\textit{table}, \textit{output\_text}}
    \State DataFrame $\gets$ \textit{table}
    % \State Initialize an empty list of relevant column names
    \State Relevant column names $\gets$ []
    \ForAll{column in the \textit{table}}
        \If{column name appears in \textit{output\_text}}
            \State Add column name to relevant column names list
        \EndIf
    \EndFor
    \If{no relevant column is found}
        \State Use all columns from \textit{table}
    \EndIf
    \State Create a new table with only relevant columns:
    \State \hspace{\algorithmicindent} Use the original table ID and title
    \State \hspace{\algorithmicindent} Use relevant column names as header
    \State \hspace{\algorithmicindent} Extract corresponding data for relevant columns
    \State \Return new decomposed table
\EndFunction

\Procedure{Main process}{}
    \State Define original\_table structure
    \State Call table\_decomposition with the original table, question, and title
    \State Call create\_decomposed\_table with the original table and output from table\_decomposition
    \State \Return the decomposed table
\EndProcedure
\end{algorithmic}
\end{algorithm}

\section{Data and Experiments}
\label{sec:DataExperiment}
We use the community standard benchmark dataset QTSUMM by \citet{zhao2023qtsumm}. Additionally, we create a decomposed dataset from the QTSUMM by removing irrelevant columns. We then train several models on both datasets where the tabular data along with the user query serve
%is the 
as input and the expected output is the summary. 

Four models are used in our experiments:
%include three models: 
T5 \cite{raffel2020exploring}, Flan-T5 \cite{chung2024scaling}, BART \cite{lewis2019bart} and OmniTab \cite{jiang2022omnitab}. To improve the summary accuracy for
%of the generated summary,
%accuracy of the 3 
the four transformer-based models,
we apply 
table decomposition, fine-tuning, and pre-processing. Next, for evaluating the query-focused table summarization task on LLMs, we utilize Llama, Mixtral, Smaug, Claude and GPT. 
% \bonnieshort{I checked this, and it's fine, so resolve red-lettering.}
We compare the results of these models and the summaries they generate against the expected summaries using BLEU \cite{papineni2002bleu}, ROUGE \cite{lin-2004-rouge}, BERTScore \cite{zhang2020bertscoreevaluatingtextgeneration} and PARENT \cite{dhingra2019handlingdivergentreferencetexts}.

%Due to resource constraints, we 
We split the test 
% \bonnieshort{We already mentioned resource constraints, so I removed that disclaimer here.}
%ing 
data into validation and test sets. Since the summary
%was 
is readily available in the test data, we 
%chose not to create it 
%choose to 
omit it 
% \bonnieshort{"choose to" is weak--it is \textbf{crucial} that it be removed, no choice, so I got rid of "choose to".}
for the validation set. Our model is then evaluated using the new validation split.
% \sangpil[]{I think we can remap to Trained Model: T5, Flan-T5 and BART, and Evaluation Model: Llama, Mixtral, Smaug, GPT3.5, Claude 2, 3, - 3.1.1 and 3.1.2 need to be changed accordingly}

\subsection{Dataset}

We randomly select 2000 training data entries from QTSUMM---including tables, summaries, and queries---to train our model. We select 500 entries as test %ing 
data and 200 entries as validation data. We train and test our model on a subset of data due to %because of 
resource limitations. The entries are selected at random ensuring that the data contains diverse table topics to ensure minimal scope of bias. We then train and evaluate our model on two versions of the QTSUMM dataset.
\begin{enumerate}
    \item Original Data: This is the original QTSUMM dataset without any preprocessing or modifications. It contains the raw tabular data, queries, and expected summaries.
    \item Decomposed Data: In this variant, we apply our table decomposition approach to the QTSUMM dataset. The tables are decomposed to retain only the most relevant columns based on the given query, removing extraneous information to address token limitations.
    % \item QA Data: For this variant, we enhance the dataset by appending fact extractions to the input data. This involves using a fact extraction model to identify and include relevant facts from the table related to the query, potentially aiding in generating more accurate summaries.
 %\sangpil[]{QA Data need to be changed into FactExtracted data}
\end{enumerate}

By training and evaluating on these two variations, we analyze the impact of our table decomposition technique and the effect of supplementing the input with extracted facts from the table.
% \bonnieshort{The red-lettered wording below is fine. But please fix these issues: (1) DO NOT create a subsubsection if there is only one subsubsection.  Just make it section 4.2. Then Fine-Tuned Models will be 4.3, and so on. (2) Orphan/widow issues, esp. in the Flattening Tables and Formatting for Fine-Tuning paragraphs.}

\subsection{Preprocessing}\label{sec:preprocessing}
Preprocessing is a crucial step before model training and evaluation. It ensures that tabular data is structured appropriately for transformer-based models. Our preprocessing pipeline includes the following steps:
\begin{itemize}
    \item \textbf{Flattening Tables:} We convert tables into a linear text format to align with sequence-based input constraints, following~\citet{hancock2019generating}, where each row is transformed into a ``key:value'' structure.
    \item \textbf{Column Selection:} To simplify the input and reduce the token count, we filter out non-relevant columns by using an LLM decomposer to break the original table into the smaller table which only contains relevant information to the query.
    \item \textbf{Tokenization:} We apply the appropriate tokenizer for each model (T5, BART, etc.), ensuring input compatibility.
    \item \textbf{Formatting for Fine-Tuning:} We append metadata like table titles and queries to ensure contextual relevance during summarization.
\end{itemize}

\subsection{Fine-tuned Models} \label{sec:flattening-tables}

We fine-tune four encoder-decoder text generation models: T5,\footnote{T5 is released under the Apache 2.0 license, users are granted a perpetual, worldwide, non-exclusive, no-charge, royalty-free, irrevocable copyright license, see here, \url{https://apache.org/licenses/LICENSE-2.0}} Flan-T5,\footnote{Flan-T5 is released under the Apache 2.0 license, users are granted a perpetual, worldwide, non-exclusive, no-charge, royalty-free, irrevocable copyright license, see above}, BART\footnote{Users of BART are granted right to redistribute and use in source and binary forms, with or without modification, see here, \url{https://github.com/mrirecon/bart/blob/master/LICENSE}} and OmniTab\footnote{OmniTab model is publically available on huggingface, see here \url{https://huggingface.co/neulab/omnitab-large}}. 
% \bonnieshort{I just realized you keep putting that apache url in footnotes--I had eliminated it from footnote 7; please consider eliminating it from others?) You only need one, and then you can keep saying: See URL above. Or something like that.}
The selection of these models is primarily driven by the need to enhance comparability with previous studies, particularly the work of \cite{zhao2023qtsumm}. By using the same models, we can draw clear comparisons between our results and theirs, providing a consistent benchmark for progress in query-focused table summarization. 

T5 is a widely used baseline for text summarization. Flan-T5 builds on T5, but it has not been as extensively used for tabular summarization as T5. We include both models to investigate improvements Flan-T5 offers over T5. Including BART allows us to explore potential improvements in tabular summarization from an alternative architecture compared to T5 and Flan-T5. OmniTab is the current state-of-the-art model on the QTSUMM dataset and utilizes a BART backbone with a pre-training setup that emphasizes tabular QA. 

It is crucial to acknowledge that these transformer-based models include restrictions regarding their context length, which may affect their capacity to manage large tables. T5 has a maximum input length of 512 tokens, whereas BART has a maximum input length of 1024 tokens. Flan-T5, being an extension of T5, has a maximum input length of 512 tokens. OmniTab, constructed on the BART architecture, maintains the 1024 token limitation. These constraints need solutions such as table decomposition to efficiently manage bigger tables and reduce information loss resulting from truncation. All these models are fine-tuned using an AMD EPYC 75F3 32-Core Processor and 3 NVIDIA A100 GPUs. We select the large versions of each model with publicly available on HuggingFace. Due to the large model and dataset sizes, we use small batch sizes during training. Additionally, we employ 4-bit quantization to reduce the memory footprint of parameter values during training while maintaining the standard precision, as demonstrated by \citet{dettmers2023qlora}. 

% \bonnieshort{Please please, do not use double backslash to force paragraph breaks!  That's not how latex works.  Just put a paragraph break like I just did. If you force it, you will not get proper first-line indentation for each paragraph. I fixed this one here, and some others below.}
For each fine-tuning experiment, we run 20 epochs. The batch size is adjusted for each model to fit into the available memory: T5 and Flan-T5 use smaller batch sizes due to more parameters, while BART and OmniTab, with fewer parameters, use larger ones. The models are fine-tuned and evaluated on two different versions of the QTSUMM dataset: the original and a decomposed version.

Before fine-tuning, we preprocess the tabular data following the steps outlined in Section~\ref{sec:preprocessing}. 
% \bonnieshort{This will become Section 4.2 \textbf{automatically} once you fix the labels and start using them correctly. I sent a slack--you are NOT supposed to \textbf{hard code} section numbers into your paragraphs.  ALL SECTIONS must use label and ref commands appropriately.  See slack.}
This ensures that the input format aligns with the transformer-based architecture constraints. Once preprocessing is complete, we apply table decomposition to reduce input size, then fine-tuning.
\subsection{Large Language Models (LLMs)}
We run experiments with various LLMs (Llama,\footnote{Meta grants a non-exclusive, worldwide, non-transferable, and royalty-free limited license for the use of Llama 3, see here, \url{https://Llama.meta.com/Llama3/license/}} Mixtral,\footnote{Mixtral is released under the Apache 2.0 license, users are granted a perpetual, worldwide, non-exclusive, no-charge, royalty-free, irrevocable copyright license, see above} GPT, Claude,\footnote{Users of Claude are granted a to deal in the Software without restriction and free of charge, see here, \url{https://github.com/Rassibassi/claude/blob/master/LICENSE}} and Smaug\footnote{Smaug is released under the Apache 2.0 license, users are granted a perpetual, worldwide, non-exclusive, no-charge, royalty-free, irrevocable copyright license, see above.})
%here, \url{https://apache.org/licenses/LICENSE-2.0}
on the same task using a zero-shot prompting approach without fine-tuning on our dataset.  
% \bonnieshort{You don't need to list the apache license twice.  I fixed the SMAUG footnote.  It had these three issues: (1) Listing apache twice is redundant; (2) listing it twice pushed the footnote into column 2, which is not desirable; (3) EVEN WORSE: By bleeding that URL into column 2, you forced column 2 to be mouse sensitive in Overleaf, so I could no longer click the PDF text and get from there to the source latex.  That's a no-no. I fixed. Resolve.} 
We employ a zero-shot prompting approach, where the models are provided with a prompt containing the table and the query, without any additional training or fine-tuning. For our experiments, %we 
%have utilized 
%use publicly %avaliable 
% available model weights available on HuggingFace. \bonnieshort{Found spelling and grammar issues in this paragraph--please check my fixes. For example, "We the prompt for each model" has a word missing; I inserted "tailor", is that the right word?}
we tailor
%used 
the prompt for each model as per the recommendations provided by the authors during their respective releases. To prepare the tabular data for input, we preprocess and flatten the tables into one-dimensional text strings. 
% This involves formatting the tables into word sequences using a ``key:value'' structure to preserve tabular information, as proposed by \citet{hancock2019generating}. Specifically, we pair each row's cells with the corresponding column headers, concatenate all rows, and then tokenize the inputs before feeding them into the models.
We followed the same process for flattening the tables as described in Section~\ref{sec:flattening-tables}.

% This allows us to train LLMs with our limited resources.
Finally, each model's performance is evaluated using BLEU, ROUGE, BERTScore and PARENT against the test dataset. This process ensures that our models not only perform well in the training
%fit the training data well
data but also generalize effectively to unseen data.

\subsection{Metrics}

% \begin{table*}
% \begin{center}
% % \caption{Results Table for Baseline Models}
% \begin{tabular}{|c|c|c|c|c|c|c|}
% \hline
% & \multicolumn{2}{|c|}{Original Data} & \multicolumn{2}{c|}{Decomposed Data} 
% % & \multicolumn{2}{c|}{QnA Data} 
% \\ 
% \hline
% & ROUGE-L & BERTScore & ROUGE-L & BERTScore 
% % & ROUGE-L & BERTScore
% \\ 
% \hline
% T5 & 0.3722 & 0.8918 & 0.3870 & 0.8949 
% % & 0.3853 & 0.9035
% \\ 
% Flan-T5& 0.3930 & 0.8974 & 0.4115 & 0.8971 
% % & 0.3852 & 0.8920
% \\
% BART & 0.4081 & 0.8949 & 0.4120 & 0.8968 
% % & 0.4088  & 0.8961
% \\
% OmniTab & 0.4405 & 0.9008 & \textbf{0.4437} & \textbf{0.9016} \\
% \hline
% \end{tabular}
% \caption{Results Table for T5, Flan-T5, BART and OmniTab with two different table handling approaches. SOTA model, REFACTOR \cite{zhao2023qtsumm} yields a ROUGE-L score of 0.422 on same task using OmniTab.}
% \label{tab:baseline}
% \end{center}
% \end{table*}

% We employ two metrics, ROUGE (Recall-Oriented Understudy for Gisting Evaluation) and BERTScore, to assess the quality and accuracy of the summaries generated by each model. We use ROUGE to emphasize recall in our evaluations, and BERTScore to overcome the limitations of n-gram overlap when considering the semantic meaning of a sequence. By considering ROUGE and BERTScore, evaluations can better reflect both the lexical accuracy and the contextual fidelity of the generated summaries.

We employ four metrics, including BLEU (Bilingual Evaluation Understudy), ROUGE (Recall-Oriented Understudy for Gisting Evaluation), BERTScore, and PARENT, to assess the quality and accuracy of the summaries generated by each model. ROUGE emphasizes recall in our evaluations, while BLEU focuses on precision by measuring the n-gram overlap between generated and reference summaries. BERTScore addresses the limitations of n-gram-based metrics by considering semantic similarity. Additionally, we incorporate PARENT, which aligns n-grams from the reference and generated texts to the underlying data before computing precision and recall. By considering BLEU, ROUGE, BERTScore, and PARENT, our evaluations provide a more comprehensive assessment of both lexical accuracy and contextual fidelity while also accounting for the alignment of generated summaries with the source data.
% \bonnieshort{I checked this, and it's fine, so resolve red-lettering.}

\section{Results and Analyses} 
\label{sec:ResultsAnalyases}

\begin{table*}[h]
\centering
\renewcommand{\arraystretch}{1.2} % Adjust row spacing
\setlength{\tabcolsep}{3pt} % Adjust column spacing
\resizebox{\textwidth}{!}{ % Resize table to fit within text width
\begin{tabular}{|c|c|c|c|c|c|c|c|c|c|c|c|c|}
\hline
 & \multicolumn{6}{c|}{Original Data} & \multicolumn{6}{c|}{Decomposed Data} \\ 
\hline
 & BLEU & ROUGE-1 & ROUGE-2 & ROUGE-L & BERTScore & PARENT & BLEU & ROUGE-1 & ROUGE-2 & ROUGE-L & BERTScore & PARENT \\ 
\hline
T5 & 0.2046 & 0.4489 & 0.2212 & 0.3722 & 0.8918 & 0.2642 & 0.2138 & 0.4544 & 0.2287 & 0.3870 & 0.8949 & 0.2787 \\ 
Flan-T5 & 0.2174 & 0.4699 & \textbf{0.2597} & 0.3930 & 0.8974 & 0.2981 & 0.2216 & 0.4851 & 0.2685 & 0.4115 & 0.8971 & 0.3124 \\ 
BART & \textbf{0.2248} & 0.4684 & 0.2312 & 0.4081 & 0.8949 & 0.3112 & 0.2405 & 0.4709 & 0.2428 & 0.4197 & 0.8968 & 0.3200 \\ 
OmniTab & 0.2213 & \textbf{0.4902} & 0.2506 & \textbf{0.4405} & \textbf{0.9008} & 0.3220 & \textbf{0.2432} & \textbf{0.4989} & \textbf{0.2756} & \textbf{0.4437} & \textbf{0.9016} & \textbf{0.3346} \\ 
\hline
\end{tabular}
} % End of resizebox
\caption{Results Table for T5, Flan-T5, BART, and OmniTab with two different table handling approaches. SOTA model, REFACTOR \cite{zhao2023qtsumm} yields a ROUGE-L score of 0.422 on the same task using OmniTab.}
\label{tab:baseline}
\vspace*{-.1in}
\end{table*}
% \sangpil[]{Please make sure what is Normal Data, Decomposed Data and QnA Data?}
% \begin{table*}
% \begin{center}
% % \caption{Results Table for Baseline Models}
% \begin{tabular}{|c|c|c|c|c|c|c|}
% \hline
% & \multicolumn{2}{|c|}{Original Data} & \multicolumn{2}{c|}{Decomposed Data} 
% % & \multicolumn{2}{c|}{QnA Data} 
% \\ 
% \hline
% & ROUGE-L & BERTScore & ROUGE-L & BERTScore 
% % & ROUGE-L & BERTScore
% \\ 
% \hline
% T5 & 0.3722 & 0.8918 & 0.3870 & 0.8949 
% % & 0.3853 & 0.9035
% \\ 
% Flan-T5& 0.3930 & 0.8974 & 0.4115 & 0.8971 
% % & 0.3852 & 0.8920
% \\
% BART & 0.4081 & 0.8949 & 0.4120 & 0.8968 
% % & 0.4088  & 0.8961
% \\
% OmniTab & 0.4405 & 0.9008 & \textbf{0.4437} & \textbf{0.9016} \\
% \hline
% \end{tabular}
% \caption{Results Table for T5, Flan-T5, BART and OmniTab with two different table handling approaches. SOTA model, REFACTOR \cite{zhao2023qtsumm} yields a ROUGE-L score of 0.422 on same task using OmniTab.}
% \label{tab:baseline}
% \end{center}
% \end{table*}

% \begin{table}
% \begin{center}
% % \caption{Results Table for LLM's}
% \begin{tabular}{|c|c|c|}
% \hline
% & \multicolumn{2}{|c|}{Original Data} \\
% \hline
% & ROUGE-L & BERTScore \\
% \hline
% Claude 2 & 0.3711 & 0.9011\\
% Claude 3 Opus& 0.3854 & 0.9022\\
% GPT-3-5 Turbo& 0.3303 & 0.8974 \\
% Llama 2-70b & 0.3543 & 0.8989\\
% Llama 3-70B & \textbf{0.4105} & \textbf{0.9103}\\
% Smaug-72B & 0.3572 & 0.9033\\
% Mixtral-8x22B & 0.3542 & 0.9035\\
% \hline
% \end{tabular}
% \caption{Results Table for LLMs.}
% \label{tab:LLMs}
% \end{center}
% % \vspace*{-.3in}
% \end{table}
The results indicate that our approach performs effectively and, in certain cases, surpasses the performance of the prior baseline technique.

The analysis, as presented in Tables~\ref{tab:baseline} and~\ref{tab:LLMs} 
reveals that Llama 3 outperforms other large language models, while OmniTab achieves the highest scores in most metrics, including ROUGE-L and PARENT. This suggests that the new Llama model offers significant advantages over earlier LLM architectures for querying tables.

In particular, the OmniTab model, when used with decomposed tabular data, emerges as the best-performing model, achieving a ROUGE-L score of \textbf{0.4437} and a PARENT score of \textbf{0.3346}. This performance surpasses the previous state-of-the-art REFACTOR model, which has a ROUGE-L score of \textbf{0.422}, also employing the OmniTab model. 
% Additionally, the ensemble model outperforms every other %individual
% model and attains a ROUGE-L score of \textbf{0.4195}. 
Additionally, it is noteworthy that models such as BART, T5, Flan-T5, and OmniTab generally exhibit slightly better performance than other LLMs. This indicates that these models possess strengths particularly suited to the task. % at hand.
 
%Moreover, 
%the results of 
Table~\ref{tab:baseline} 
%indicate 
shows a consistent trend across all models,
with 
%of having 
better performance on the 
%table decomposition data. 
decomposed data. This suggests that 
breaking down
%decomposing 
the data into its constituent parts, before presenting it to the models, enhances the models' ability to generate accurate summaries. This improvement occurs because the models can focus more effectively on %the
relevant information when it is presented in a structured and decomposed format. This structured approach allows the models to concentrate on essential elements, thereby improving the overall accuracy and quality of the generated summaries.
% \bonnieshort{You can earn back a lot of space by not making a single column table into a full table across the page--and you can use smaller font for tables. Also, boldface highest score for every single measure.}

% \bonnieshort{Is everything you promised here?  In your response, you stated "T-test analysis revealed ..." but I don't see it. Note that this was considered a weak evaluation, so it's best to follow through on promises.}

% Despite the overall improvement seen with the table decomposition process, there are instances where the models fail to properly summarize the table. This failure occurs when key columns, which should have been included in the summarization process, are removed during the data decomposition stage. In such cases, we manually replace those instances without decomposition and rerun the tests through our model. In this hybrid table decomposition process, we achieve a ROUGE-L score of \textbf{0.4385}. This highlights the scope of improvement and a limitation of the decomposition approach, indicating that careful consideration must be given to the selection and presentation of data to ensure optimal performance of the models. 

\begin{table}[h!]
\centering
\renewcommand{\arraystretch}{1.2} % Adjust row height for better readability
\setlength{\tabcolsep}{4pt} % Reduce column spacing
\resizebox{\linewidth}{!}{%
\begin{tabular}{|l|c|c|c|c|c|c|}
\hline
Model & BLEU & ROUGE-1 & ROUGE-2 & ROUGE-L & BERTScore & PARENT \\ 
\hline
Claude 2 & 0.2238 & 0.4816 & 0.2464 & 0.3702 & 0.9011 & 0.2673 \\ 
Claude 3 Opus & 0.2334 & \textbf{0.4975} & 0.2561 & 0.3857 & 0.9022 & 0.2843 \\ 
GPT-3.5 Turbo & 0.2303 & 0.4593 & 0.2255 & 0.3301 & 0.8974 & 0.2548 \\ 
Llama 2-70B & 0.2134 & 0.4694 & 0.2435 & 0.3543 & 0.8989 & 0.2457 \\ 
Llama 3-70B & \textbf{0.2539} & 0.4948 & \textbf{0.2649} & \textbf{0.4105} & \textbf{0.9103} & \textbf{0.2953} \\ 
Smaug-72B & 0.2230 & 0.4801 & 0.2333 & 0.3369 & 0.9033 & 0.2407 \\ 
Mixtral-8x22B & 0.2198 & 0.4790 & 0.2412 & 0.3542 & 0.9035 & 0.2476 \\ 
\hline
\end{tabular}%
}
\caption{Results Table for LLMs.}
\label{tab:LLMs}
\vspace*{-0.2in}
\end{table}

\section{Human Evaluation Study and Error Analysis}

Although automatic n-gram overlap metrics like ROUGE are valuable, they have limitations in evaluating semantically similar text. A qualitative analysis is necessary to better understand any system's strengths and limitations. Therefore, we conduct an error analysis and a human evaluation study to gain additional insights into our system. %Additionally, we perform human evaluation of table %decomposition to assess how well the table %decomposition itself performs.

% \bonnieshort{Widow/orphan alert.}

\subsection{Evaluation of Table Decomposition}

To provide a qualitative perspective of the tabular decomposition method, we perform a human evaluation study. We employ a Likert scale ranging from 1 to 5 to assess the completeness and accuracy of the information in the decomposed tables. 
 
\begin{itemize}
    \itemsep0em
    \item A score of 1 means that most or all relevant information is missing
    \item A score of 2 suggests that some relevant information is present, though the table remains largely incomplete. 
    \item A score of 3 indicates that most relevant information is available, but not comprehensive. %all. 
    \item A score of 4 %indicates 
    signifies that all relevant information is present, but some may still be missing, or the table includes irrelevant details. %decomposition includes non relevant details that is not required. 
    \item A perfect score of 5 reflects that the table contains only the relevant information. 
\end{itemize}
% \vspace*{-.2in}
We randomly sample 100 data points, evaluate them, and present the results. We choose 100 samples due to time and resource constraints. For table decomposition, we score 4.46 on a scale of 5, indicating there is room for improvement.

% \justin[]{Do you think this needs a table @Sangpil also this definitely needs some elaboration. We }

\subsection{Human Evaluation of Table Summarization}
For the human evaluation study, three MS-level non-algorithm developers independently evaluate summaries generated by multiple models based on accuracy, relevance, and clarity, using a five-point rating scale. The results of human evaluation of some of our best models are shown in Table~\ref{tab:humanevaluation}. Llama3 is the best-performing model in terms of accuracy. We can also conclude that the Omnitab model performs better for tabular data, as evidenced by its improved Rouge-L and accuracy scores compared to the BART model.

% Models       Accuracy   Relevance   Clarity

% Bart          3.22         3.92        4.67
% Omnitab       3.41         4.12        4.58
% Llama3         
\begin{table}[h!]
\centering
\begin{tabular}{|c|c|c|c|}
\hline
\textbf{Models} & \textbf{Accuracy} & \textbf{Relevance} & \textbf{Clarity} \\ \hline
BART & 3.22 & 3.92 & 4.67 \\ \hline
Omnitab & 3.41 & 4.12 & 4.58 \\ \hline
Llama3 & 3.77 & 4.08 & 4.42 \\ \hline
\end{tabular}
\caption{Human Evaluation Results. Bart and Omnitab model evaluation are done on decomposed data. Llama3 evaluation is done on original data.}
\label{tab:humanevaluation}
\vspace*{-.1in}
\end{table}
% \vspace*{-.1in}

To measure inter-rater agreement, we select a common set of randomly generated 100 summaries, which are independently rated by all three evaluators. We then compute the Intraclass Correlation Coefficient (ICC)~\cite{koo2016icc} to assess the consistency of the ratings across the three criteria. The ICC score obtained is 0.7768, indicating good level agreement among the evaluators.

\subsection{Error Analysis for Table Summarization}	
For the error analysis, we categorize errors into factual incorrectness, irrelevant information, hallucinations, and repetition \cite{zhao2023qtsumm}. Irrelevant information includes cases where the facts extracted by the model are correct but are not pertinent to the user query. We manually review 100 randomly-selected samples of generated summaries and evaluate them as errors within these categories. This analysis reveals which stages in our process fail, along with the limitations of our current approach and areas for future improvement. The results of error analysis for the Omnitab model using our table decomposition approach, are presented in Table~\ref{tab:error}.

% Table Structure
% Error                      Total Counts
% Factual incorrectness       x.x
% Irrelevant information       x.x
% hallucination               x.x
% Repetition                  x.x
% Correct                     x.x
% \begin{table}[h!]
% \centering
% \begin{tabular}{|{\raggedright\arraybackslash}p{5cm}}
% \hline
% \textbf{Error} & \textbf{Total Counts} \\ \hline
% Factual incorrectness &  18 \\ \hline
% Irrelevant information & 14 \\ \hline
% Hallucination & 6 \\ \hline
% Repetition & 2 \\ \hline
% Correct & 60 \\ \hline
% \end{tabular}
% \caption{Total count for different errors.}
% \label{tab:error}
% \end{table}

\begin{table}[h!]
\centering
\begin{tabular}{|c|c|}
\hline
\textbf{Error} & \textbf{Total Counts} \\ \hline
Factual incorrectness &  18 \\ \hline
Irrelevant information & 14 \\ \hline
Hallucination & 6 \\ \hline
Repetition & 2 \\ \hline
Correct & 60 \\ \hline
\end{tabular}
\caption{Total number of errors in different categories on 100 random examples}
\label{tab:error}
\vspace*{-.1in}  
\end{table}   

As shown in Table~\ref{tab:error}, according to human evaluators, 60\% of the summaries generated are correct. Factual incorrectness is the most common, which accounts for 18\% of total samples, followed by irrelevant information which accounts for 14\% of all the samples. Hallucinations and repetitions are less frequent with 6\% and 2\% probability respectively. This analysis highlights the areas where the summarization model requires improvement, particularly in addressing the factual correctness and relevance of the facts generated. By focusing on these aspects, future iterations of the model can enhance overall performance and reliability. 

\section{Conclusion and Future Work}
\label{sec:conclusion}
This study presents a novel query-focused approach to tabular summarization, integrating table decomposition with advanced text generation models (T5, Flan-T5, BART, and Omnitab). We mitigate token limitations of existing models by efficiently handling large and complex tables, thereby improving upon the current state-of-the-art REFACTOR.

Our best-performing model 
%attain an 
attains a ROUGE-L score of 0.4437, setting a new high score for query-focused tabular summarization. By exploring diverse models (T5, Flan-T5, BART, Omnitab, Llama, Mixtral, GPT),
%and 
we gain valuable insights into their capabilities 
%when it comes to 
on this task.

Future research directions include exploring additional models and LLMs, such as Flan-T5 XXL, a larger variant than the one used in this study. Improving ensemble techniques by training models on different types of data or queries is another future direction. For example, some models could be trained on simple queries and others on complex queries, or one model could train on the full dataset while others train on the decomposed dataset. This would allow the ensemble model to leverage the respective strengths of individual models.

Moreover,  Llama3 demonstrates the highest accuracy in query-based table summarization tasks, as demonstrated by human evaluation, reflecting superior fact extraction capabilities, but OmniTab achieves the highest ROUGE score, indicating stronger overall summarization performance. This result suggests that while Llama3 excels in accuracy and detailed fact retrieval, OmniTab provides a more coherent and comprehensive summary. To leverage the strengths of both models, future research could explore the development of a hybrid model that combines Llama3's precise fact extraction with OmniTab's robust table summarization capabilities. Such an approach could potentially enhance both the accuracy and overall quality of table summarization, offering a more balanced and effective solution.

% \bonnieshort{I split up the next two paragraphs. The red-letter text is otherwise fine. But I'm not sure you've included everything you promised, e.g., BARTScore was going to be mentioned, ...}

In scenarios where even a decomposed table exceeds the summarization model's token capacity, additional strategies become necessary. One approach is further decomposition—breaking the table into even smaller, manageable segments. Alternatively, hierarchical methods can process large tables by first summarizing individual sections and then combining these summaries into a cohesive final output. Techniques such as iterative summarization or chunked processing may also help preserve key information when handling extremely large datasets. Evaluating the effectiveness of these approaches is an important direction for future research to enhance model robustness in managing complex, large-scale tabular data.

Additionally, implementing continuous learning mechanisms %can 
may allow 
%the 
models to evolve in response to new data and user feedback without requiring full retraining. This adaptive learning 
%approach 
may help maintain the relevance and efficiency of the summarization system \cite{wu2024continual}.

Another promising avenue for future work is to explore evaluation metrics beyond BLEU, ROUGE, BERTScore and PARENT such as BARTScore. It utilizes a pre-trained BART model and has demonstrated robust efficiency in evaluating the quality of output summaries by analyzing both lexical overlap and semantic similarity. By applying this, we could assure a more thorough assessment of model performance and produced summaries.
% PARENT is especially designed for assessing table-to-text generation jobs, including both content selection and surface realization. 
% Integrating this indicator into our evaluation approach will enhance its alignment with the complexities of query-focused table summarizing, hence assuring a more thorough assessment of model performance and produced summaries.

By employing these enhancements, we can solidify the utility of our research and push the boundaries of what automated systems can achieve in the realm of intelligent data summarization. 

%This interdisciplinary study has deepened our understanding of NLP and summarization tasks, advancing intelligent summarization systems that enable accurate, relevant, and context-aware summaries tailored to user needs.

%While promising, our approach has limitations. We suggest future research directions: exploring advanced language models, implementing query decomposition for complex queries, dynamically decomposing datasets based on query types, enhancing ensemble techniques through specialized model training, adopting multi-aspect evaluation frameworks, and incorporating continuous learning mechanisms.

%Our interdisciplinary collaboration has deepened our understanding of NLP and summarization tasks. By addressing challenges of large tabular data, our research advances intelligent summarization systems, enabling accurate, relevant, context-aware summaries tailored to user needs.

% \sangpil[]{I think this last pargraph need to be improved, including our results, and contribution.}
% Through meticulous experiments and evaluations using automated metrics such as ROUGE and BERTScore, we demonstrate efficacy of our approach. However, we acknowledge the limitations of this study and have suggest some potential future research. Our collaborative efforts have fostered interdisciplinary collaboration and propelled us toward a deeper understanding of NLP and summarization tasks.


\section*{Limitations}
%We come across s
% \bonnie{Since Appendix A.1 is cited multiple times, perhaps you want to say "see Appendix A.1, example X" ?}
Several
%A number of 
shortcomings 
have emerged during this research study. First, %ly, 
%in most cases, 
our models tend to perform %the 
best on 
simple queries or recall queries that ask for direct information from the table (see Appendix~\ref{sec:example-1}, Example 1). However, our model demonstrates reduced performance when handling queries that involve complex reasoning across multiple columns or require identifying intricate relationships and patterns within the table (see Appendix~ \ref{sec:example-2}, Example 2). This limitation arises from the model’s difficulty in comprehending and reasoning across several data points within a single query, leading to challenges such as factual inaccuracies or hallucinations. Specifically, our method struggles with accurately synthesizing summaries when queries involve intricate temporal, causal, or relational patterns, as these exceed the scope of the model’s current decomposition approach. 

To address the challenges of complex queries, we propose integrating a ``chain-of-thought'' reasoning process in future model iterations. This approach would involve decomposing complex queries into a sequence of logical, incremental reasoning steps. By breaking down the task into smaller units, the model could tackle one part of the query at a time, gradually building up to a complete and accurate summary. This decomposition would enhance the model’s ability to reason over multifaceted data relationships and reduce the likelihood of hallucinations. Additionally, we aim to explore ensemble methods, using standard decomposition for simpler tasks and a hierarchical multi-step approach for complex reasoning.
%To mitigate this, 
% we plan to enhance our approach by integrating more advanced models and techniques that improve reasoning capabilities. For example, a future research could explore a hybrid table decomposition method that combines both raw and decomposed data, allowing the models to maintain access to critical information needed for complex queries and improving their ability to identify patterns and trends.

% \sangpil[]{It is better not to refer appendix. The work is only refer anything inside main text.}
% \sangpil[]{Each every paragraphs in limitations, need to be saying how we remediate our limitation. Not just our weakness of the study  Addressed - Xinlei}
Second, while %through 
table decomposition is intended
%we aim 
to filter out noise for improving model accuracy, it can
%. However, in some cases, table decomposition loses 
sometimes lose
%the 
important data or 
%other specific data 
specific information needed for recall and comparison queries (Appendix~\ref{sec:example-3}, Example 3). In 
such cases, % the case above %case 
our model may hallucinate
%hallucinates and generates 
and generate facts not present in the table. Table decomposition also negatively impacts queries that require an overarching understanding of
%any overarching query for 
trends or patterns, as these advanced queries 
often need
%require 
more information for accurate summaries, and %sometimes the 
table decomposition
%drops this additional information.
can omit this additional data.
To mitigate this issue, we plan to
%We plan to mitigate this issue by implementing 
implement a dynamic decomposition strategy that %. This strategy 
adjusts the extent of decomposition based on the complexity of the query. For simple recall queries, more aggressive decomposition is applied, whereas, for complex queries, a lighter decomposition is used to retain more information, in accordance with previous work \cite{ye2023large}. 
% Additionally, we plan to introduce a verification step where the model's outputs are cross-checked against the original table to prevent hallucinations.

When the original table exceeds the sequence length of the LLM-decomposer, our current approach cannot process the entire table directly. In such cases, there is a possibility of truncation of the input and potentially degradation of table decomposition quality. This limitation highlights the need for further investigation into alternative strategies. For instance, hierarchical decomposition or table segmentation can divide large tables into smaller, manageable parts, ensuring the integrity of the summarized content through improved token allocation. Additionally, exploring large language models with higher context windows also accommodates larger inputs without sacrificing performance.
% \bonnieshort{I reviewed and this red-letter paragraph is fine, although I really do not see much of the promised wording  in this paper, including in the Limitations.}

% Furthermore, when the original table exceeds the LLM-decomposer's sequence length, our technique struggles to process the entire table, resulting in input truncation and perhaps lower decomposition quality. This constraint highlights the need for more study into approaches that can manage large tables while maintaining decomposition integrity. Exploring strategies such as table segmentation, which divides huge tables into smaller bits, or building LLM designs with higher sequence length capabilities, might improve our approach's efficiency and transparency.

Finally, the human evaluation and error analysis in this study have been
%were 
conducted by only three evaluators,
%\bonnieshort{What does "reviewer" mean?  Do you mean "researcher"? Do you mean "author"? (I think "reviewer" will be confusing to the "reviewers" :-) ) }
which may potentially introduce human biases. With a limited number of evaluators, there is an increased risk that %the reviewer's 
individual perspectives, experiences, and subjective interpretations could disproportionately influence the results. This may lead to certain errors being overlooked or specific patterns being misinterpreted, potentially 
%thus might affect 
affecting the reliability of the human evaluation. A more diverse group of evaluators could provide a broader range of insights, helping to ensure that the analysis is comprehensive and more representative of a general consensus. To mitigate this limitation, we aim to involve a larger and more diversified panel of expert evaluators or employing additional measures such as cross-validation and consensus discussions to reduce the impact of individual biases on the evaluation outcomes.

% Finally,cwhile BERTScore and ROUGE-L ratings offer valuable insights into the quality of summaries, we acknowledge the limitations of relying solely on these metrics. Due to the timeline and nature of this research, we are not able to apply human measurements or metrics for the model predictions. Thus, we are constrained to using ROUGE-L and BERTScore, which are not always optimal measures of performance for language models. To mitigate this limitation, another future area for research is a multi-aspect evaluation framework that goes beyond lexical similarity. This framework includes measuring the correctness of the information, the degree to which the summary addresses key points and the logical flow and organization of ideas within the summary.
% Additionally,incorporating
% human evaluations in future research could 
% further validate the model's performance and ensure that the generated summaries meet user expectations.

\section*{Ethical Considerations}

In conducting this research on query-focused tabular summarization, several ethical considerations %were 
are central.
%paramount. 
%Additionally, the 
The potential for bias in data and model outputs is critically assessed, with our efforts to use
%. Efforts are made to utilize 
a diverse dataset,
%to minimize bias, 
although it is acknowledged that the complete elimination of bias is challenging and ongoing efforts are necessary.

The environmental impact of training and deploying large language models (LLMs) is another significant consideration. Techniques such as 4-bit quantization are employed to reduce computational resources and mitigate the carbon footprint associated with extensive model training \cite{dettmers2023qlora}. Furthermore, the risk of model inaccuracies, including the generation of erroneous facts, is recognized. This underscores the importance of continuous monitoring and iterative improvement of models to enhance accuracy and reliability \cite{zhao2023qtsumm}.
By mitigating these ethical considerations, this research aims to responsibly advance the field of query-focused table summarization while mitigating potential negative impacts.

This paper is assisted by AI software solely for formatting and grammatical checking purposes. No underlying
%important 
ideas or content are generated by AI. All original ideas, analysis, and conclusions in this paper are solely created by the listed authors.

\bibliography{MoodMenders}

\appendix
% \makeindex
\section{Appendix}

We provide examples of query, expected summary, and model outputs with decomposing table process.

\begin{center}
\label{sec:example-1}
\centering
\textbf{{EXAMPLE 1}}

% \newline
\underline{Non-Decomposed Table} 

List of Columbo episodes -Season-4 (1974–75)
\includegraphics[width=1\linewidth]{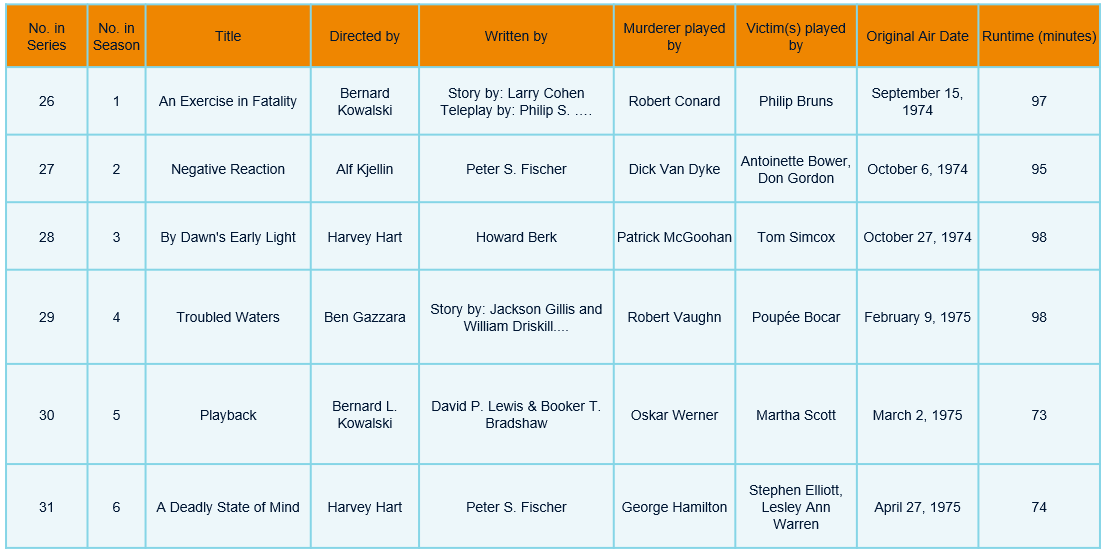}

\underline{Decomposed Table}
\includegraphics[width=1\linewidth]{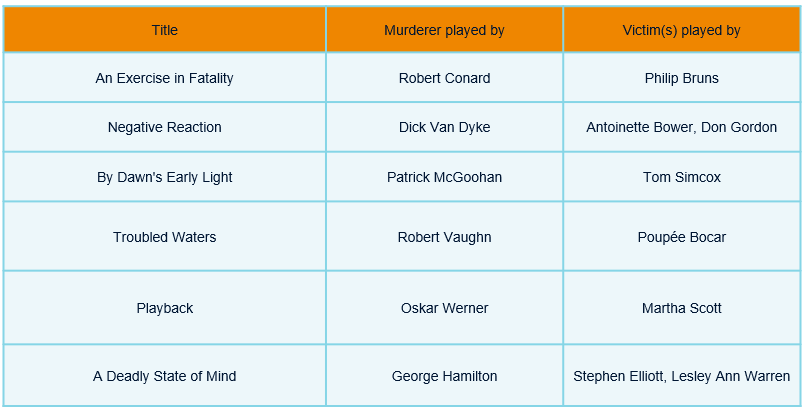}

\label{fig:ensemble}
\end{center}

\textbf{Query} - Who played the role of the murderer in the episode "Negative Reaction", and who were the victims in this particular episode?

\textbf{Expected Summary} - In the episode "Negative Reaction," Dick Van Dyke plays the role of the murderer. The victims in this special episode are Antoinette Bower and Don Gordon.

\textbf{Returned Summary (BART)} - In the episode ``Negative Reaction,'' Dick Van Dyke plays the role of the murderer. The victims in this episode are Antoinette Bower and Don Gordon

\textbf{ROUGE-L Score}- 0.8444

\begin{center}
\label{sec:example-2}
\centering
\textbf{{EXAMPLE 2}}

% \newline
\underline{Non-Decomposed Table}

S and DJR 7F 2-8-0-Construction

\includegraphics[width=1\linewidth]{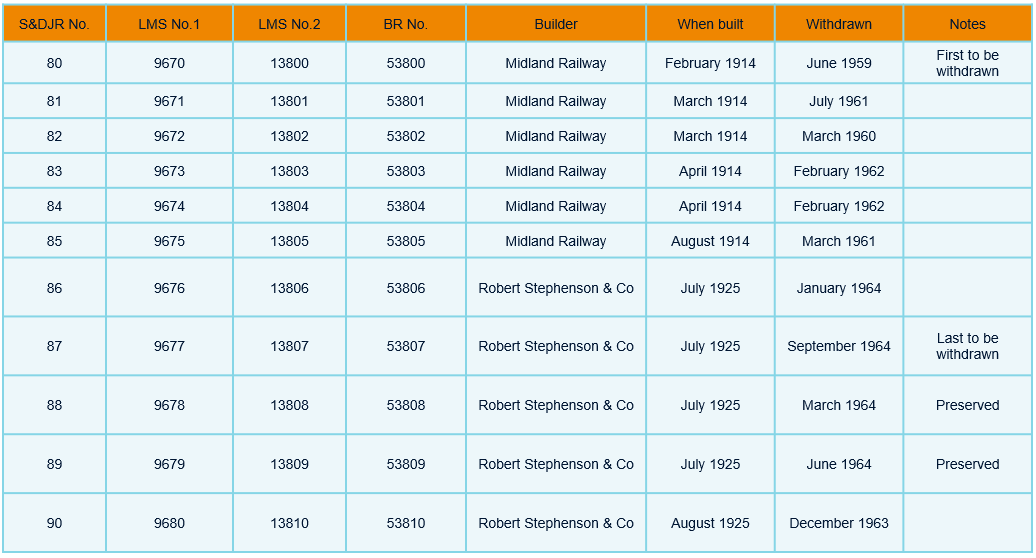}

\underline{Decomposed Table}
\includegraphics[width=1\linewidth]{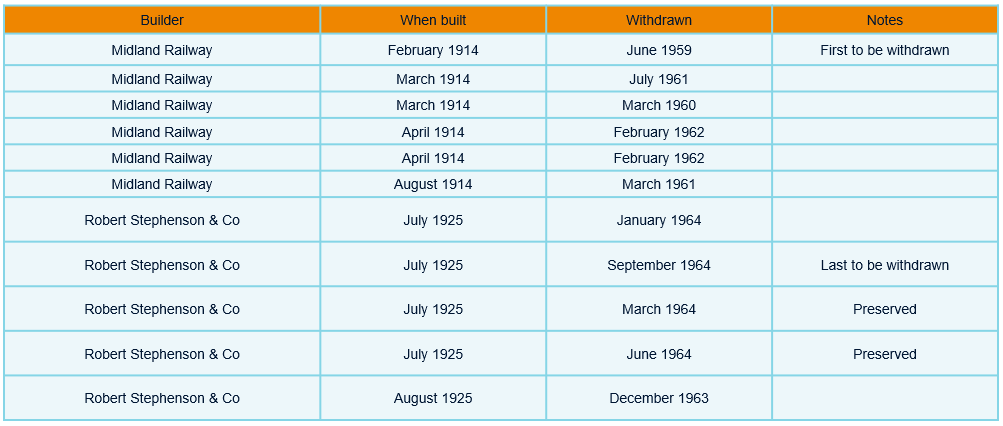}

\label{fig:ensemble}
\end{center}

\textbf{Query} - How does the timeline of withdrawal correlate with the dates of construction for the locomotives listed in the table, especially those built by Robert Stephenson \& Co in 1925?

\textbf{Expected Summary} - The time to withdraw for trains in table look like has order, old one withdraw first then new one. For trains made by Midland Railway between February and August 1914, their withdraw time is from June 1959 to February 1962. The first made train is first to be withdraw. Not same, trains made by Robert Stephenson \& Co in 1925 have withdraw time between December 1963 and September 1964. This shows these trains, because they are new, withdraw later than those made by Midland Railway. This way tells that withdraw time has relation with make time, as old trains often took out from use before new ones.

\textbf{Returned Summary (BART)} - The timeline of withdrawal for the locomotives listed in the table seems to have a strong correlation with the dates of construction for the locomotives listed in the table. The locomotives built by Robert Stephenson \& Co in 1925 were all completed between January 1964 and September 1964. This suggests that the timeline of withdrawal for these locomotives was not as long as the dates of construction for the other locomotives listed in the table.

\textbf{ROUGE-L Score- 0.2111}
% \hline

\begin{center}
\label{sec:example-3}
\centering
\textbf{{EXAMPLE 3}}

% \newline
\underline{Non-decomposed Table} 

List of international cricket centuries at Rose Bowl - One Day International centuries 
\includegraphics[width=1\linewidth]{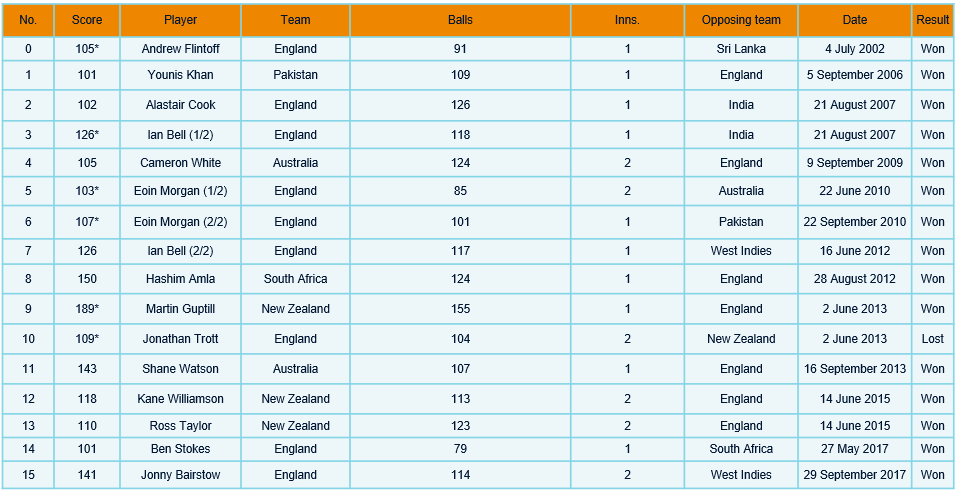}

\underline{Decomposed Table}
\includegraphics[width=1\linewidth]{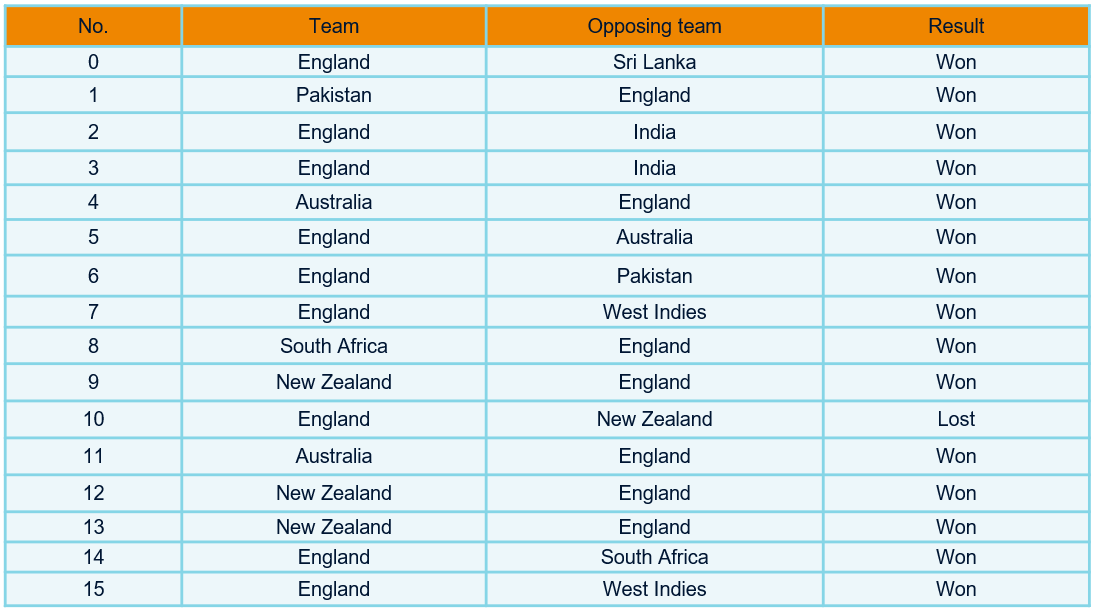}

\label{fig:ensemble}
\end{center}

\textbf{Query} - How did team's innings (Inns.) affect their chances of winning or losing the match?

\textbf{Expected Summary} - From table, we can see that team bat first (Inns. 1) win match in 10 out of 16 times, and team bat second (Inns. 2) win in 6 out of 16 times. This maybe means bat first can give team advantage, because they can set high target for other team to chase.

\textbf{Returned Summary (BART)} - In 1948-1949 Ashes series, teams' innings (Inns.) have big effect on their chances to win or lose match. When England play first innings, they have more chance to win, if they score 200 runs or more, and when they score 150 runs or less, they can win match. For example, when England score 400 runs in first innings against Sri Lanka, they win match, but when they bat second innings, their chances of win go down. In same way, if England score 200 or less runs in second innings against Pakistan and India, they get more chance of win, because they score more runs, make match more likely to end in win.

\textbf{ROUGE-L Score}- 0.1454
% \hline

\end{document}